\documentclass{article}

    \PassOptionsToPackage{numbers, compress}{natbib}
    \usepackage[preprint]{neurips_2025}

\usepackage[utf8]{inputenc} % allow utf-8 input
\usepackage[T1]{fontenc}    % use 8-bit T1 fonts
\usepackage{hyperref}       % hyperlinks
\usepackage{url}            % simple URL typesetting
\usepackage{booktabs}       % professional-quality tables
\usepackage{amsfonts}       % blackboard math symbols
\usepackage{nicefrac}       % compact symbols for 1/2, etc.
\usepackage{microtype}      % microtypography
\usepackage{xcolor}         % colors

\usepackage{amsmath}
\usepackage[pdftex]{graphicx}
\usepackage{colortbl}
\usepackage{multirow}
\usepackage{caption}

\usepackage[capitalize]{cleveref}
\crefname{section}{Sec.}{Secs.}
\Crefname{section}{Section}{Sections}
\Crefname{table}{Table}{Tables}
\crefname{table}{Tab.}{Tabs.}

\title{Recollection from Pensieve: Novel View Synthesis via Learning from Uncalibrated Videos}

\author{%
  Ruoyu Wang
    \\
  Transcengram\\
  \texttt{dwawayu@gmail.com} \\
  \And
  Yi Ma \\
  Transcengram \\
  The University of Hong Kong \\
  \texttt{mayi@hku.hk} \\
  \And
  Shenghua Gao \\
  Transcengram \\
  The University of Hong Kong \\
  \texttt{gaosh@hku.hk} \\
}

\begin{document}

\maketitle

\begin{abstract}
Currently almost all state-of-the-art novel view synthesis and reconstruction models rely on calibrated cameras or additional geometric priors for training. These prerequisites significantly limit their applicability to massive uncalibrated data. To alleviate this requirement and unlock the potential for self-supervised training on large-scale uncalibrated videos, we propose a novel two-stage strategy to train a view synthesis model from only raw video frames or multi-view images, without providing camera parameters or other priors.
In the first stage, we learn to reconstruct the scene implicitly in a latent space without relying on any explicit 3D representation. Specifically, we predict per-frame latent camera and scene context features, and employ a view synthesis model as a proxy for explicit rendering. This pretraining stage substantially reduces the optimization complexity and encourages the network to learn the underlying 3D consistency in a self-supervised manner. The learned latent camera and implicit scene representation have a large gap compared with the real 3D world. To reduce this gap, we introduce the second stage training by explicitly predicting 3D Gaussian primitives. We additionally apply explicit Gaussian Splatting rendering loss and depth projection loss to align the learned latent representations with physically grounded 3D geometry. In this way, Stage 1 provides a strong initialization and Stage 2 enforces 3D consistency - the two stages are complementary and mutually beneficial. Extensive experiments demonstrate the effectiveness of our approach, achieving high-quality novel view synthesis and accurate camera pose estimation, compared to methods that employ supervision with calibration, pose, or depth information.
The code is available at \url{https://github.com/Dwawayu/Pensieve}.

%Our method eliminates the need for camera calibration, pre-trained models, or geometric priors, making it feasible to train directly on large-scale uncalibrated video datasets. 

%We propose a novel two-stage training strategy to train the view synthesis model without any camera parameters, relying solely on uncalibrated video data. The two stages are complementary and mutually beneficial.
%In the first stage, we reconstruct the scene implicitly in a latent space without relying on any explicit 3D representation. Specifically, we predict per-frame latent cameras and context frame features, and employ the view synthesis model as a proxy for explicit rendering. This pretraining stage substantially reduces optimization complexity and encourages the network to learn underlying 3D consistency in a self-supervised manner.
%In the second stage, we additionally predict explicit 3D Gaussian primitives to bridge the gap between the pretrained latent space and the real 3D world. We apply explicit rendering and projection losses, aligning the learned latent representations with physically grounded 3D geometry.
%Our method eliminates the need for camera calibration, pre-trained models, or geometric priors, making it feasible to train directly on large-scale uncalibrated video datasets. Extensive experiments demonstrate the effectiveness of our approach, achieving high-quality novel view synthesis and accurate camera pose estimation.
\end{abstract}

\vspace{-0.5\baselineskip}
\section{Introduction}
\label{sec:introduction}
\vspace{-0.5\baselineskip}
Simultaneously reconstructing the scene and localizing the camera is a long-standing and challenging task in computer vision. Solving this task has the potential to enable the training of fundamental 3D vision networks on large-scale, uncalibrated video data. Previous large reconstruction models typically rely on preprocessed camera parameters and point clouds obtained via SfM (Structure-from-Motion) or SLAM (Simultaneous Localization and Mapping). However, such preprocessing can be time-consuming, require densely sampled views~\cite{colmap}, or depend on additional 3D information~\cite{densergbdslam}, which limits their applicability in more challenging and unconstrained datasets.

Several recent approaches jointly predict scene and camera parameters with neural networks, training them from scratch on video data. The key challenge lies in differentiably establishing correspondences across views. Earlier depth-based (or point cloud-based) methods~\cite{sfmlearner, monodepth1} achieve this by employing bilinear interpolation, allowing each projected point to receive gradients from its four neighboring pixels. More recent Gaussian Splatting-based methods~\cite{selfsplat, hong2024pf3plat} render Gaussian primitives and propagate gradients from pixels within the extent defined by the Gaussian scale. However, these methods still restrict the differentiable matching to a limited spatial neighborhood. Moreover, their carefully designed 3D representations often introduce optimization biases, which lead the network to converge to suboptimal solutions.

In this paper, we propose a two-stage training strategy that combines implicit reconstruction pretraining with explicit reconstruction alignment. Given an input video, our model estimates camera parameters of each frame and predicts context features for a strict subset of those frames. In the pretraining stage, we employ a view synthesis model (inspired by LVSM~\cite{lvsm}) to re-render (i.e., predict) all the input frames themselves for self-supervision. This fully end-to-end architecture avoids the challenges associated with explicit representations and enables implicit scene reconstruction in a self-supervised manner.

However, due to the lack of explicit 3D consistency in pretraining, the learned implicit reconstruction can diverge from the actual physical 3D structure. In essence, the stage1 model behaves like an autoencoder for the target image, with camera parameters acting as an intermediate representation. This means it can only interpolate views at latent cameras, rather than synthesizing views from given real cameras. To address this limitation, we introduce a second training stage to align the implicit reconstruction with the real 3D geometry. This stage additionally predicts explicit 3D Gaussian primitives and computes Gaussian Splatting rendering loss~\cite{3dgs} and depth reprojection loss~\cite{sfmlearner, monodepth1}. This alignment enables both novel view synthesis and accurate camera estimation in an entirely self-supervised setting. Experimental results demonstrate that the two training stages are mutually beneficial: the stage1 pretraining accelerates convergence and improves performance, while the stage2 alignment forces the network to learn the true 3D structure and camera parameters.

In summary, our contributions are as follows:
\vspace{-0.5\baselineskip}
\begin{itemize}
\item We propose a self-supervised training framework for novel view synthesis (NVS) networks that rely solely on uncalibrated videos, achieving high-quality NVS and accurate camera pose estimation.
\item We introduce implicit reconstruction pretraining to address the limitations of prior self-supervised 3D reconstruction methods that rely on explicit representations.
\item We propose explicit reconstruction alignment to enforce 3D consistency in the pretrained network, aligning its latent space with the real space of the scene.
\item We also introduce an interpolated frame enhanced prediction scheme to address the issue of insufficient camera alignment when only two input views are available.
\end{itemize}
\vspace{-0.5\baselineskip}

\section{Related Works}
\paragraph{Novel View Synthesis.}
Novel view synthesis and 3D reconstruction are fundamental computer vision tasks that have been extensively studied, especially since the advent of neural radiance fields (NeRF)~\cite{nerf}. Traditional approaches such as Structure-from-Motion (SfM)~\cite{buildingrome, colmap} reconstruct sparse point clouds and estimate camera parameters. Building upon this, new scene representations like NeRF~\cite{nerf} and 3D Gaussian Splatting (3DGS)~\cite{3dgs}, have been proposed to further improve the quality of NVS. Moreover, many subsequent improvements have been proposed to enhance its rendering quality~\cite{mipnerf, mipsplatting}, geometric accuracy~\cite{li2023neuralangelo, 2dgs}, memory efficiency~\cite{yang2024spectrally, lu2024scaffold}, and reconstruction speed~\cite{chen2022tensorf, ingp}, as well as to extend it to dynamic scenes~\cite{park2021nerfies, nerfsceneflow, kplanes, yu2024cogs, huang2024scgs}.

These representations can be made generalizable by incorporating neural networks. Some methods leverage neural networks to predict generalizable radiance fields~\cite{yu2021pixelnerf, chen2021mvsnerf, li2021mine, tian2023mononerf}, while others directly predict Gaussian primitives to reconstruct the scene from sparse views in a single feed-forward pass~\cite{pixelsplat, mvsplat}. Beyond architectures based on cost volumes~\cite{chen2021mvsnerf} or epipolar lines~\cite{suhail2022light, wang2021ibrnet, pixelsplat, mvsplat}, several methods adopt fully data-driven approaches, which benefit more from large-scale datasets~\cite{szymanowicz2024splatter, tang2024lgm, xu2024grm, gslrm, lvsm}. Among them, LVSM~\cite{lvsm} proposes to directly predict the target frame using a neural network, instead of relying on explicit rendering, thus eliminating the inductive biases introduced by explicit 3D representations. However, these methods require training on datasets with known camera parameters, which limits their applicability to larger-scale, uncalibrated video data.
\vspace{-\baselineskip}
\paragraph{Camera-free Novel View Synthesis.}
Estimating camera parameters using SfM is not always reliable, especially under sparse-view settings or in scenes with large textureless regions. To address this, several methods have proposed jointly optimizing cameras with NeRF~\cite{nerfmm, barf, localrf,scnerf, truong2023sparf} or 3DGS~\cite{colmapfree3dgs, jiang2024construct, monogs} during per-scene reconstruction, and leveraging pretrained networks to further improve performance~\cite{noponerf, park2024splinegs}.

Several methods attempt to train generalizable networks that jointly estimate camera parameters and reconstruct scenes. The difficulty of this setting varies depending on the type of supervision provided during training. Flare~\cite{zhang2025flare} uses ground-truth point clouds and camera parameters for supervision, while NoPoSplat~\cite{noposplat} and VicaSplat~\cite{li2025vicasplat} are trained with given camera parameters. These methods benefit from either direct or indirect camera supervision, thereby alleviating the difficulty of optimization. Splatt3R~\cite{smart2024splatt3r} leverages the pretrained large-scale reconstruction network~\cite{wang2024dust3r, mast3r} to estimate both point clouds and camera poses, while PF3plat~\cite{hong2024pf3plat} incorporates pretrained depth estimation and matching networks. CoPoNeRF~\cite{coponerf} and GGRt~\cite{li2024ggrt} utilize pretrained feature extractors and provide pose to supervised the matching. FlowCam~\cite{smith2023flowcam} employs pretrained optical flow to indirectly supervise both camera estimation and reconstruction. In contrast, SelfSplat~\cite{selfsplat} proposes to jointly optimize a camera network and a 3DGS network from uncalibrated video data without pretrained priors, while provide camera intrinsics to simplify the task. Our method requires only uncalibrated video frames or multi-view images, without relying on any additional data or pretrained priors, achieving high-quality NVS and accurate camera pose estimation, thereby unlocking the potential for training on large-scale and more diverse datasets.

\section{Method}

\subsection{Definition}
Given an uncalibrated video with a length of \(N\) frames $\{\textbf{I}_i\ |\ i=1,...,N\}$, our network \(\mathcal{M}_{\theta}\) predicts, for each frame $\textbf{I}_i$, the corresponding context features \(\mathbf{F}^c_i\), pixel-aligned Gaussian primitives $\mathbf{G}_i$, and camera intrinsic $\mathbf{K}_i$ and extrinsic $\mathbf{P}_i$. Mathematically, this can be formulated as follows:
\begin{equation}
\{\mathbf{F}^c_i, \mathbf{G}_i, \mathbf{K}_i, \mathbf{P}_i | i=1, ..., N\} = \mathcal{M}_{\theta}(\{\mathbf{I}_i | i=1, ..., N\}),
  \label{eq:model}
\end{equation}
where the prediction for each frame is conditioned on all frames, allowing the model to incorporate multi-view context in its estimation.

Thus, given a target camera $\{\mathbf{K}_t, \mathbf{P}_t\}$, we enable a unified model $\mathcal{R}^M$ to synthesize the corresponding target view image $\hat{\mathbf{I}}^M_t$, similar to LVSM~\cite{lvsm}.
In addition, we can leverage Gaussian Splatting Rasterization~\cite{3dgs} to render the target view image $\hat{\mathbf{I}}^G_t$.
Formally, this can be expressed as:
\begin{equation}
\hat{\mathbf{I}}^M_t = \mathcal{R}^{M}(\mathbf{K}_t, \mathbf{P}_t, \mathbf{F}^c_{1:N}, \mathbf{K}_{1:N}, \mathbf{P}_{1:N}) \hspace{0.5cm} \hat{\mathbf{I}}^G_t = \mathcal{R}^{G}(\mathbf{K}_t, \mathbf{P}_t, \mathbf{G}_{1:N})
  \label{eq:render_lvsm}
\end{equation}
% In addition, we can leverage Gaussian Splatting Rasterization~\cite{3dgs} to render the target view image $\hat{\mathbf{I}}^G_t$, formulated as:
% \begin{equation}
% \hat{\mathbf{I}}^G_t = \mathcal{R}^{G}(\mathbf{K}_t, \mathbf{P}_t, \mathbf{G}_{1:N}).
%   \label{eq:render_gs}
% \end{equation}
For the definition of the camera, we assume an ideal pinhole camera model with the principal point at the image center. The network predicts the unknown focal lengths $f_x$ and $f_y$. For the extrinsics $\mathbf{P}$, the network directly predicts the rotation as a quaternion $\mathbf{R}^q \in \mathbb{R}^4$, along with the translation $\mathbf{t} \in \mathbb{R}^3$. We assume that all frames in a given video share the same intrinsic parameters, which is reasonable for most video sequences. To achieve this, we simply average the predicted intrinsic parameters across all frames.

For each Gaussian primitive $\mathbf{G} = \{\boldsymbol{\mu}, \alpha, \mathbf{q}, \mathbf{s}, \mathbf{c}\}$, we adopt the geometrically accurate 2D Gaussian Splatting formulation proposed in~\cite{2dgs}. Consequently, our network is required to predict the following parameters for each primitive: the center position $\boldsymbol{\mu} \in \mathbb{R}^3$, the opacity $\alpha \in \mathbb{R}$, the rotation represented as a quaternion $\mathbf{q} \in \mathbb{R}^4$, the anisotropic scale $\mathbf{s} \in \mathbb{R}^2$, and the color modeled via spherical harmonics (SH) coefficients $\mathbf{c} \in \mathbb{R}^k$, where $k$ denotes the number of SH coefficients used to represent view-dependent appearance.

\subsection{Network Architecture}

\begin{figure}
  \centering
  \includegraphics[width=\linewidth]{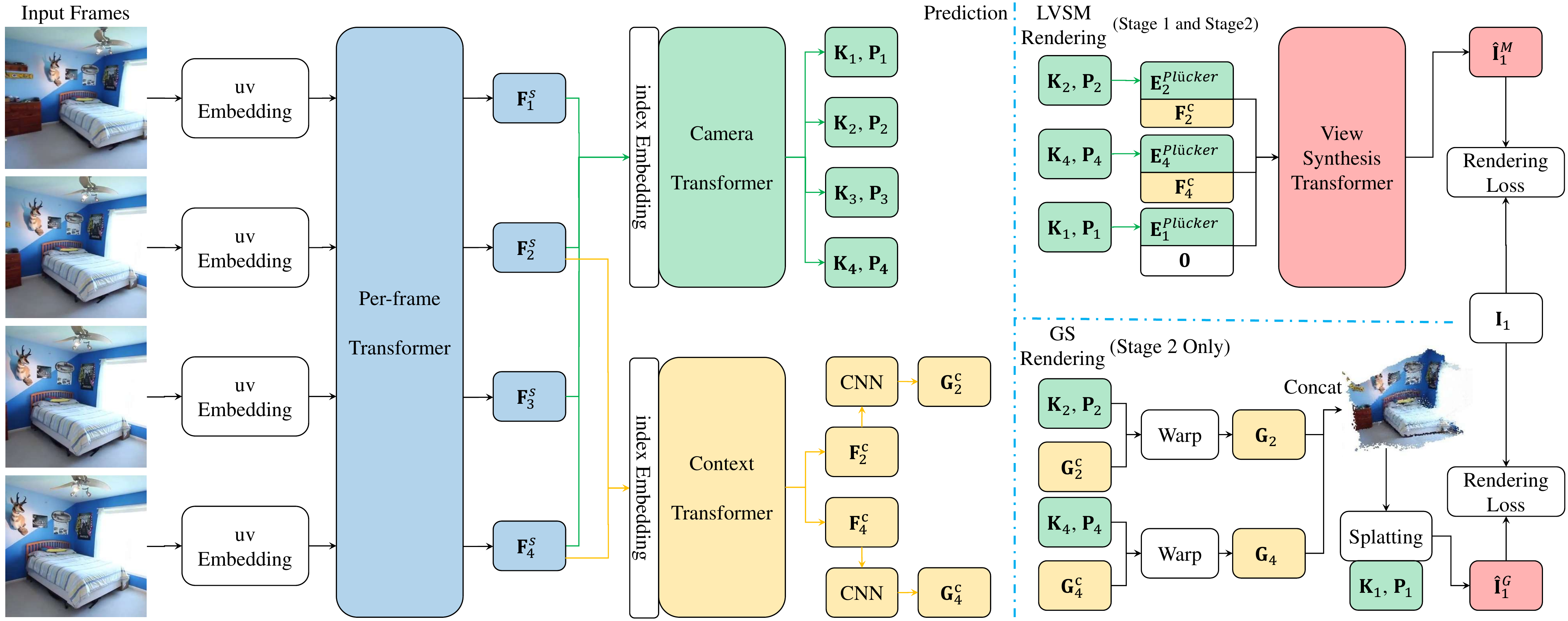}
  \caption{\textbf{Pipeline}. 
  Given an $N$-frame uncalibrated video (with $N = 4$ shown), our method first extracts per-frame features $\mathbf{F}^s$ using a shared-weight Per-Frame Transformer. Features from all frames are passed through a Camera Transformer to predict per-frame camera parameters $\mathbf{K}$ and $\mathbf{P}$. A sampled proper subset of features (e.g., $\{ \mathbf{F}^s_2, \mathbf{F}^s_4 \}$) is fed into the Context Transformer to predict the corresponding context features $\mathbf{F}^c$ and pixel-aligned Gaussian primitives $\mathbf{G}^c$ in the camera coordinate system. Subsequently, we rerender each frame and compute the rendering loss, as illustrated by the example of the first frame. For LVSM rendering, we apply Plücker coordinates embedding to represent camera features, and use the View Synthesis Transformer to predict the target image $\hat{\mathbf{I}}^M_1$. In Stage 2, we warp the Gaussian primitives into world coordinates, and concatenate them to additionally render the target image $\hat{\mathbf{I}}^G_1$ explicitly using the predicted camera parameters.}
  \label{fig:pipeline}
  \vspace{-\baselineskip}
\end{figure}

Although our model \(\mathcal{M}_{\theta}\) is designed to output per-frame predictions as formulated in~\cref{eq:model}, it is important to note that during training, the context frames must form a proper subset of the frames used for camera prediction, as we will discuss in~\cref{sec:pretraining}. Therefore, our model should be structured into two components: the Camera Transformer, which predicts the camera parameters, and the Context Transformer, which reconstructs the scene (implicitly and explicitly), with the constraint that the input to the Context Transformer is always a strict subset of the frames fed to the Camera Transformer during training. Naturally, since both components require processing image features, we design a shared Per-Frame Transformer to extract per-frame image features in advance, reducing the overall parameters. We also implement the LVSM module in~\cref{eq:render_lvsm} using the same architecture. As shown in~\cref{fig:pipeline}, all four transformers in our pipeline share the same architecture type, similar to GS-LRM~\cite{gslrm}, but differ in embedding methods, number of layers, and hidden dimensions.
\vspace{-\baselineskip}
\paragraph{Transformer.} All of our transformers transform the input maps into output maps of the same resolution. In each transformer, we first patchify the input maps and flatten each patch into tokens. Following GS-LRM~\cite{gslrm}, each transformer is composed of stacked consecutive blocks, where each block can be described as:
\begin{equation}
\mathbf{T}_{l}^m = \text{SelfAttn}(\text{LN}(\mathbf{T}_{l-1})) + \mathbf{T}_{l-1} \hspace{0.5cm} \mathbf{T}_{l} = \text{MLP}(\text{LN}(\mathbf{T}_{l}^m)) + \mathbf{T}_{l}^m,
\end{equation}
where $\mathbf{T}$ denotes the tokens at each layer, $l$ indicates the current layer index, SelfAttn refers to the self-attention module, MLP is a multi-layer perceptron, and LN stands for layer normalization. After passing through $L$ transformer blocks, we unpatchify $\mathbf{T}_L$ to reconstruct output maps of the same resolution as the input. We now describe the differences among the four Transformers.

For the \textbf{Per-Frame Transformer}, we feed each frame independently into the network to avoid any leakage of inter-frame information. We simply use the pixel coordinates as the positional embedding. Specifically, each input map is augmented with two additional channels whose values are the normalized pixel coordinates $(u / W, v / H)$, where $(u, v)$ denotes the coordinates, and $(W, H)$ represents the resolution of the map.

For the \textbf{Camera Transformer}, we treat all image features as a group of input maps. For each map in this group, we simply use its index as the positional embedding. Specifically, each input map is augmented with one additional channel whose value is $i / N$, where $i$ denotes the index of the current map within the group, and $N$ is the total number of maps. We simply apply Global Average Pooling (GAP) over each output map, followed by a lightweight MLP to predict the camera parameters.

For the \textbf{Context Transformer}, we randomly sample a strict subset of the image features as input. We similarly apply index embedding to the input. The predicted context features $\mathbf{F}^c$ are directly fed into a lightweight CNN to predict pixel-aligned Gaussian primitives $\mathbf{G}^c$ in the camera coordinate system.

For the \textbf{View Synthesis Transformer}, we follow the approach of LVSM~\cite{lvsm} by representing all cameras using Plücker coordinates embedding. We apply the Plücker coordinates embedding of the corresponding camera to the context features as an implicit reconstruction of the scene. For the target camera, we concatenate its embedding with a zero map $\mathbf{0}$ to form the target tokens. Finally, we take only the updated target tokens from the output as the synthesized target view image.

\subsection{Latent Reconstruction Pretraining}
\label{sec:pretraining}
Training networks to estimate camera parameters and scene structure solely from uncalibrated videos or multi-view images has long been a fundamental challenge in computer vision. The key is differentiably identifying correspondences between images. Early depth-based~\cite{sfmlearner, monodepth2} and plane-based~\cite{falnet, planedepth} methods employed bilinear interpolation to make the matching differentiable, but each projected point only receives gradients from four neighboring pixels. Gaussian Splatting-based approaches~\cite{selfsplat} have leveraged rendered Gaussian primitives to establish correspondences, yet each point is still only affected by the pixels within its Gaussian scale range. These limitations increase the optimization difficulty and often trap the network in suboptimal solutions. LVSM~\cite{lvsm} addresses this by introducing an end-to-end network that renders (predicts) the target image so that each target pixel is differentiable to all input pixels, effectively circumventing the aforementioned limitations of explicit 3D representations. Motivated by this, we propose using LVSM~\cite{lvsm} as the renderer for network pretraining, enabling the model to implicitly learn better correspondences.

Unlike explicit reconstruction, LVSM does not enforce 3D consistency on the gradients propagated to the camera parameters. As a result, the implicit reconstruction behaves more like an encoding–decoding process, with the camera parameters serving as an intermediate representation. In general, after reconstructing a scene, we can query an image using any given camera. Therefore, the degrees of freedom (DoF) of the camera parameters correspond to the minimal DoF required to encode the image. Our pretraining follows this principle: it encodes the target image into a latent representation with the same DoF as the camera parameters and queries the target image from the context images, even though this latent representation does not necessarily align with the real camera. We treat the latent representation directly as a camera and further transform it into the Plücker embedding. In our latent space, the features of context images enable the scene reconstruction, while the target camera is the middle representation to encode and decode the target image.

However, there is one critical exception that can break this encoding mechanism—when the target image itself is included among the context images. In this case, the network can trivially encode the target image by simply referencing its index within the context set, bypassing any meaningful reconstruction. To prevent this, we ensure throughout our training process that the context images are always a strict subset of the input video frames, so that at least one target image lies outside the context set.

\Cref{fig:pipeline} illustrates the pipeline of our approach. We encode each input frame into camera parameters, randomly sample context frames and predict their features, and embed these context features with their corresponding camera parameters to form an implicit scene reconstruction. Given the camera of the first frame as the target, we use the View Synthesis Transformer to decode the first frame. Although our predicted cameras may not align with the real 3D world, we still treat them as valid camera parameters and adopt the Plücker embedding as their features. After decoding all input frames sequentially, we compute the rendering loss as a combination of an MSE and an LPIPS loss:
\begin{equation}
\mathcal{L}^M_{\text{render}} =\frac{1}{\sum w}\sum^N_{i=1} w_i (\text{MSE}(\hat{\mathbf{I}}^M_i, \mathbf{I}_i) + \lambda_{1} \, \text{LPIPS}(\hat{\mathbf{I}}^M_i, \mathbf{I}_i))
\hspace{0.5cm} 
w_i=\begin{cases}
1, &\mathbf{I}_i \notin \text{context set} \\
w_\text{low}, &\mathbf{I}_i \in \text{context set}
\end{cases},
\label{eq:loss_render_m}
\end{equation}
where we assign a lower weight (e.g., $w_\text{low} = 0.1$) to the loss of reconstructed context images to mitigate the trivial encoding issue.

\subsection{Explicit Reconstruction Alignment}
Although the implicit reconstruction pretraining avoids the optimization challenges associated with explicit representations, it inherently lacks explicit 3D consistency constraints. As a result, the reconstructed space may diverge from the true physical world. While the cameras used during pretraining could be physically correct, it is more likely that the network leverages the same DoF to encode unintelligible latent features. Therefore, additional explicit 3D alignment is required to enforce 3D consistency.

Similar to SelfSplat~\cite{selfsplat}, we choose to use the Gaussian Splatting rendering loss together with a depth projection loss to align self-supervisely. We use the latent pretrained weight and additionally predict pixel-aligned Gaussian primitives from the context features using a lightweight CNN, as illustrated in~\cref{fig:pipeline}. The center position $\boldsymbol{\mu}$ of each Gaussian primitive is obtained by first predicting a depth map and then back-projecting it into 3D space, formulated as:
\begin{equation}
\boldsymbol{\mu}_i^{u,v} = \mathbf{R}_i D_i^{u,v} \mathbf{K}_i^{-1} [u \ v \ 1]^\top + \mathbf{t}_i,
\label{eq:center_position}
\end{equation}
where $D_i^{u,v}$ is the predicted depth at pixel $(u, v)$, $\mathbf{K}_i$ and $\mathbf{R}_i, \mathbf{t}_i$ are the intrinsic and extrinsic camera parameters for frame $i$, respectively. Importantly, the depth map $D_i$ is also predicted by the CNN trained from scratch.

In the stage 2 training, we retain the pretraining loss~\cref{eq:loss_render_m} while additionally introducing the Gaussian Splatting rendering loss $\mathcal{L}^G_{\text{render}}$. Specifically, we concatenate all Gaussian primitives predicted from the context features and rerender the entire input video $\{\hat{\mathbf{I}}^G_i \mid i = 1, \dots, N\}$. The Gaussian Splatting rendering loss $\mathcal{L}^G_{\text{render}}$ follows the same formulation as~\cref{eq:loss_render_m}, except that $\hat{\mathbf{I}}^M_i$ is replaced by $\hat{\mathbf{I}}^G_i$.

For predicted depth $\mathbf{D}$ by our model and the rendered depth $\hat{\mathbf{D}}$ by the Gaussian Splatting, we compute the projection loss~\cite{sfmlearner} and the edge-aware smoothness loss~\cite{monodepth1}. Given the depth map for the $i$-th frame, we use the camera parameters to compute the per-pixel projection onto the $j$-th frame. We then synthesize the $i$-th frame $\hat{\mathbf{I}}^D_i$ by bilinearly sampling colors from the $j$-th frame and compute the MSE loss between the synthesized and ground-truth images. The projection and smoothness losses are:
\begin{equation}
\mathcal{L}_{\text{proj}} = \text{MSE}(\hat{\mathbf{I}}^D_i, \mathbf{I}_i)
\hspace{0.5cm}
\mathcal{L}_{\text{ds}} = |\partial_x \mathbf{D}_i \odot e^{-\gamma \|\partial_x \mathbf{I}_i\|_1}| + |\partial_y \mathbf{D}_i \odot e^{-\gamma \|\partial_y \mathbf{I}_i\|_1}|,
\label{eq:loss_depth}
\end{equation}
where \(\gamma\) controls the smoothness around edges~\cite{falnet}. We apply the same loss to the rendered depth $\hat{\mathbf{D}}$, simply replacing $\mathbf{D}$ with $\hat{\mathbf{D}}$ in the~\cref{eq:loss_depth}. It is important to note that each input frame has an associated rendered depth map $\hat{\mathbf{D}}$, whereas only the context frames have corresponding predicted depth maps $\mathbf{D}$. For each projection, the frame $j$ is randomly sampled from the set of input frames.

Therefore, the final loss for our stage 2 training is:
\begin{equation}
\mathcal{L}_{\text{stage2}} = \mathcal{L}^M_{\text{render}} + \mathcal{L}^G_{\text{render}} + \lambda_1 \mathcal{L}_{\text{proj}} + \lambda_2\mathcal{L}_{\text{ds}},
\end{equation}
which is averaged over pixels, views, and batches. $\lambda_{1}$ and $\lambda_{2}$ are hyperparameters used to balance different loss terms. Notably, $\lambda_{1}$ is gradually reduced to zero during training, as the projection loss is affected by the occlusion problem~\cite{monodepth2}.

\subsection{Interpolated Frame Enhanced Prediction}
\label{sec:if}
During training, we randomly sample the length of the input video, with a minimum of two frames, and shuffle the input frame order to simulate the multi-view reconstruction task. However, when only two frames are provided as input, the context frame must be a strict subset of the input frames, leaving only one frame available as the context. In this case, there are large unseen regions when rendering the other frame. Gaussian Splatting produces large holes when rendering these unseen regions, causing the network to predict oversized Gaussian primitives and underestimate camera motion in an attempt to fill the holes. As a result, alignment performance degrades under two-frame input conditions.

To address this issue, after completing the two-stage training, we apply a specific inference-time strategy tailored for the two-frame case. At inference, we interpolate the two input frames into a three-frame video and re-feed it into the network to obtain the final output. This interpolation is performed using our LVSM rendering. Specifically, after the initial two-frame prediction, we average the predicted camera parameters of the two input frames to generate an intermediate camera, which serves as the target camera. Using LVSM rendering in~\cref{eq:render_lvsm}, we render the interpolated middle frame, which is then combined with the original two frames to form a three-frame video that is re-input into the network for standard prediction.
\section{Experiments}
\begin{table}[t]
\centering

\caption{Qualitative Comparison on RealEstate10K~\cite{re10k}. The best is in \textbf{bold} and the second is \underline{underlined} in each metric. \textbf{K}, \textbf{P}, \textbf{D}, and \textbf{M} denote additional training data of camera intrinsics, camera extrinsics, depth, and matching, respectively. The results show that our method achieves the best NVS quality. Moreover, our method still attains comparable camera pose estimation performance even without additional supervision from extrinsics or matching. *: stands for our method with optimized camera pose of test view.}
\label{tab:exp_re10k}
\resizebox{1\linewidth}{!}{
\begin{tabular}{cccccccccc}
\toprule
Method & Training Data & PSNR↑ & SSIM↑ & LPIPS↓ & RRA@5↑ & RRA@15↑ & RTA@5↑ & RTA@15↑ \\ 

\hline

\rowcolor{lightgray} \multicolumn{9}{c}{RealEstate10K~\cite{re10k} Target-aware Evaluation} \\

PF3plat & Video + $\mathbf{KDM}$ & 22.84 & 0.790 & 0.190 & \textbf{92.8} & \textbf{98.3} & \textbf{58.6} & \textbf{91.5} \\

SelfSplat & Video + $\mathbf{K}$ & 22.04 & 0.772 & 0.237 & 70.9 & 92.1 & 37.1 & 67.0 \\

Ours     & Video        & \textbf{26.53} & \textbf{0.843} & \textbf{0.115} & \underline{84.8} & \underline{98.2} & \underline{48.1} & \underline{85.2}  \\

\hline

\rowcolor{lightgray} \multicolumn{9}{c}{RealEstate10K~\cite{re10k} Target-aligned Evaluation} \\

CoPoNeRF & Video + $\mathbf{KP}$  & 19.66 & 0.668 & 0.327 & \underline{89.8} & \textbf{98.4} & \underline{47.0} & \textbf{85.8} \\

PF3plat & Video + $\mathbf{KDM}$ & 20.04 & 0.643 & 0.260 & \textbf{92.0} & \underline{97.7} & 37.8 & \underline{85.6} \\

SelfSplat & Video + $\mathbf{K}$  & 18.50 & 0.598 & 0.347 & 60.1 & 85.1 & 24.1 & 49.4 \\

Ours     & Video        & \underline{22.20} & \underline{0.712} & \underline{0.176} & 85.5 & \underline{97.7} & \textbf{50.3} & 85.3  \\

Ours*     & Video        & \textbf{23.96} & \textbf{0.778} & \textbf{0.145} & 85.5 & \underline{97.7} & \textbf{50.3} & 85.3  \\

\bottomrule

\end{tabular}
}
\vspace{-\baselineskip}
\end{table}
\begin{figure}
  \centering
  \includegraphics[width=\linewidth]{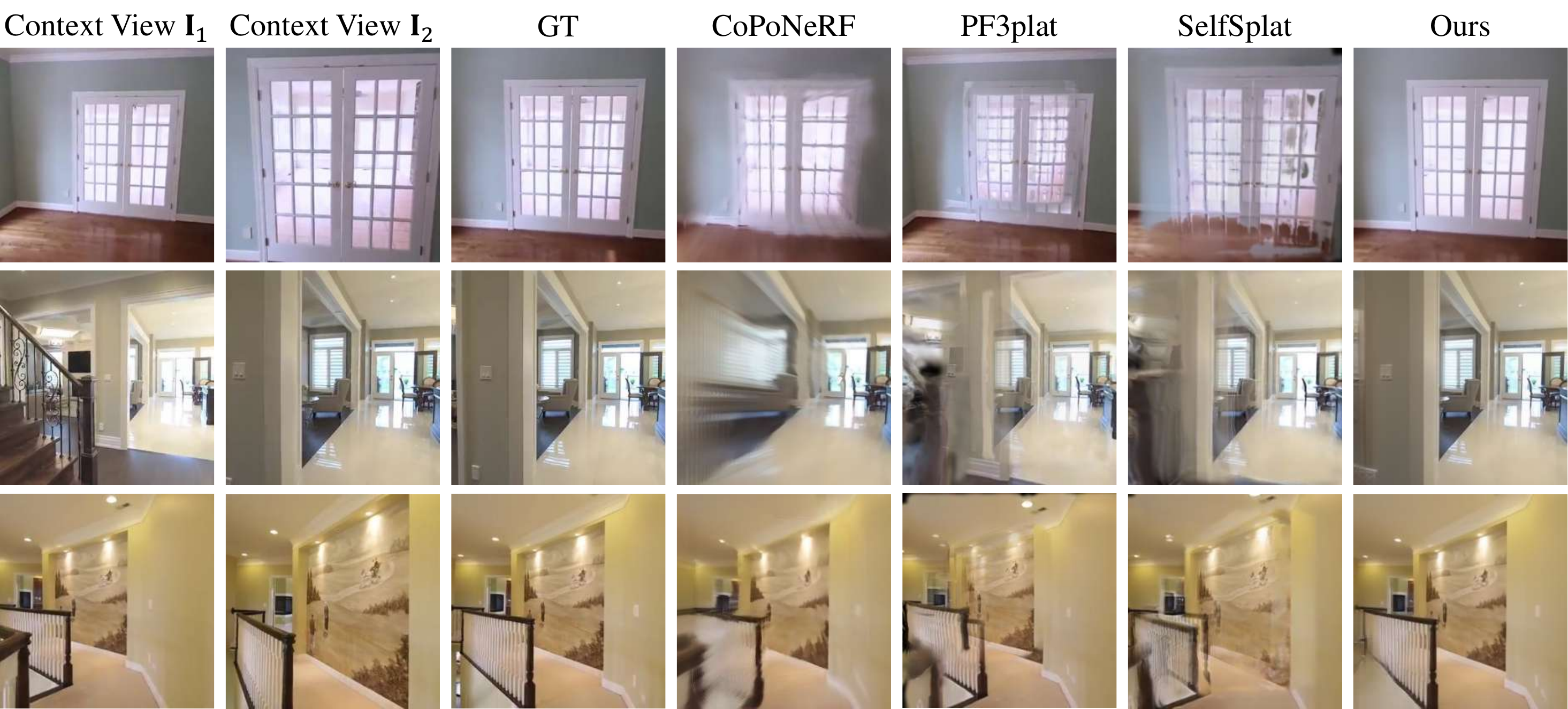}
  \caption{Novel view synthesis comparisons on RealEstate10K~\cite{re10k}. Our method better aligns with ground-truth poses and image content.}
  \label{fig:qualitative}
\end{figure}

\subsection{Datasets}
\paragraph{RealEstate10K.} The RealEstate10K~\cite{re10k} dataset contains 80,000 clips extracted from 10,000 YouTube videos. We follow the train-test split used in PixelSplat~\cite{pixelsplat}. We use only the video frames as training data for our model and ignore the other preprocessed data.
\vspace{-\baselineskip}
\paragraph{DL3DV-10K.} The DL3DV-10K~\cite{ling2024dl3dv} dataset contains 10,510 videos spanning both bounded and unbounded scenes. Following the protocol of prior work~\cite{selfsplat}, we finetune all the models on DL3DV using the weights trained on RealEstate10K~\cite{re10k}. Evaluation is performed on the DL3DV-140 benchmark, using the small-overlap and large-overlap splits proposed by PF3plat~\cite{hong2024pf3plat}.

\subsection{Implementation Details}
All of our Transformers adopt an architecture similar to GS-LRM~\cite{gslrm}, with a patch size of $8 \times 8$. The Per-frame Transformer consists of 12 blocks with a hidden dimension of 768. The Camera Transformer, Context Transformer, and View Synthesis Transformer each consist of 8 blocks with a hidden dimension of 512. Our network also includes several simple shortcut connections. Please refer to the supplementary material for more details.

For RealEstate10K, We train at a resolution of $256\times  256$, using video clips of random lengths with $N \in [2, 7]$. For DL3DV-10K, we train at a resolution of $256\times 448$, sampling video frames with a random length $N \in [2, 5]$. We shuffle the input frame order during training.

\subsection{Evaluation Protocols}
For all comparison methods, we follow their original settings and provide the additional data for training and evaluation to ensure fairness. Our evaluation protocols also follow their settings, using two frames as context frames and selected intermediate frames as test frames (specified in~\cite{pixelsplat, hong2024pf3plat}). Based on how the camera pose of the test view is obtained, we categorize the evaluation into Target-aware Evaluation~\cite{hong2024pf3plat, selfsplat} and Target-aligned Evaluation~\cite{coponerf}. As a contrast, our method is trained on variable-length, unordered video frames, without any fine-tuning tailored to such inputs or additional data.

\textbf{Target-aware Evaluation}. Following prior works~\cite{hong2024pf3plat, selfsplat}, we predict the camera of the test view using both the context images and the test image as inputs to the trained camera pose network. This setting provides greater overlap between views for pose estimation. However, as discussed in~\cref{sec:pretraining}, this acutally encodes and decodes the target view, rather than performing true novel view synthesis.

\textbf{Target-aligned Evaluation}. Following~\cite{coponerf}, only context views are given as input to the camera network to estimate pose. The ground-truth camera parameters of the test view are then aligned to the estimated context camera poses for rendering. This ensures a more realistic evaluation of the NVS capabilities. Since our input views are sparse and no prior information is introduced, the reconstruction may not align with the ground truth, leading to discrepancies in the aligned test-view poses. Therefore, following previous works~\cite{barf,nerfmm}, we freeze all parameters except the test-view extrinsics during evaluation and apply the Gaussian Splatting rendering loss for 40 optimization iterations to further refine the alignment. We present the results with both optimized and non-optimized camera poses for comparison.

We evaluate the novel view synthesis performance using PSNR, SSIM, and LPIPS metrics, and use the Relative Rotation Angle (RRA) and Relative Translation Angle (RTA) to evaluate the predicted poses between pairs of input views. The @5 and @15 metrics indicate the percentage of view pairs with angular errors within 5° and 15°, respectively.

\subsection{Performance Comparisons}

\begin{table}[t]
\centering

\caption{Qualitative Comparison on DL3DV-140~\cite{ling2024dl3dv}. The best is in \textbf{bold} in each metric. The results demonstrate that our method performs the best in rendering quality and pose accuracy. PF3plat requires the video to have camera intrinsics and depth, and use a pretrained matching network to estimate initial pose and extract features. SelfSplat's inputs contain both camera intrinsics and video. Our method only takes uncalibrated videos as input. *: stands for our method with optimized camera pose of the test view.}
\label{tab:exp_dl3dv}

\resizebox{1\linewidth}{!}{
\begin{tabular}{ccccccccccc}
\toprule

\multirow{2}{*}{Method} & \multicolumn{5}{|c}{small} & \multicolumn{5}{|c}{large} \\
 & \multicolumn{1}{|c}{PSNR↑} & SSIM↑ & LPIPS↓ & RRA@5↑ & RTA@5↑ & \multicolumn{1}{|c}{PSNR↑} & SSIM↑ & LPIPS↓ & RRA@5↑  & RTA@5↑  \\ 

\rowcolor{lightgray} \multicolumn{11}{c}{DL3DV-Benchmark~\cite{ling2024dl3dv} Target-aware Evaluation} \\

PF3plat & 18.77 & 0.583 & 0.291 & 71.9 &  39.6 & 21.35 & 0.679 & 0.223 & 89.9 & 46.0 \\

SelfSplat & 19.11 & 0.586 & 0.396 & 74.1 & 46.0 & 21.70 & 0.689 & 0.309 & 89.9 & 58.9 \\

Ours & \textbf{21.77} & \textbf{0.662} & \textbf{0.183} & \textbf{84.9} & \textbf{58.3} & \textbf{23.89} & \textbf{0.750} & \textbf{0.133} & \textbf{93.5} & \textbf{59.0} \\

\rowcolor{lightgray} \multicolumn{11}{c}{DL3DV-Benchmark~\cite{ling2024dl3dv} Target-aligned Evaluation} \\

PF3plat & 17.01 & 0.466 & 0.345 & 76.3 & 33.1 & 19.80 & 0.591 & 0.257 & 90.7 & 44.6 \\

SelfSplat & 16.99 & 0.478 & 0.451 & 61.9 & 36.7 & 20.32 & 0.617 & 0.340 & 88.5 & 48.9 \\

Ours & 19.36 & 0.543 & 0.242 & \textbf{82.7} & \textbf{55.4} & 22.21 & 0.673 & 0.164 & \textbf{92.8} & \textbf{61.9}  \\

Ours* & \textbf{20.02} & \textbf{0.587} & \textbf{0.222} & \textbf{82.7} & \textbf{55.4} & \textbf{22.52} & \textbf{0.696} & \textbf{0.153} & \textbf{92.8} & \textbf{61.9} \\

\bottomrule

\end{tabular}
}

\end{table}
The quantitative results on RealEstate10K~\cite{re10k} shown in~\cref{tab:exp_re10k} demonstrate that, although our method is trained only on raw video frames, it achieves the best novel view synthesis quality, with slightly lower camera pose estimation accuracy. This is because CoPoNeRF~\cite{coponerf} benefits from additionally provided camera extrinsics as supervision for pose estimation, while PF3plat~\cite{hong2024pf3plat} leverages a pretrained matching network~\cite{lindenberger2023lightglue} that supplies rich correspondence information. The qualitative results in~\cref{fig:qualitative} further show that our method accurately aligns the target pose and synthesizes the image, even from two input frames with large textureless regions.

The quantitative results on DL3DV~\cite{ling2024dl3dv} shown in~\cref{tab:exp_dl3dv} indicate that our method achieves the best performance in both novel view synthesis and camera pose estimation. This is attributed to the increased difficulty of the DL3DV dataset, where PF3plat~\cite{hong2024pf3plat} is constrained by the accuracy of its pretrained matching network~\cite{lindenberger2023lightglue}, which may fail for pose estimation, and SelfSplat~\cite{selfsplat} suffers from the optimization limitations of explicit Gaussian primitives. In contrast, our method benefits from implicit reconstruction to learn latent correspondences through a fully end-to-end network, leading to superior performance. Even without any camera information as guidance, by introducing the Stage 2, our method still achieves the best performance for camera pose estimation.

% Comparison results on RealEstate10K~\cite{re10k} Despite training solely on raw video frames, our method outperforms all baselines. The only exception is a slightly lower pose accuracy under the RealEstate10K Target-aware Evaluation. This is largely attributed to the use of pretrained matching~\cite{lindenberger2023lightglue} in PF3plat and ground-truth camera poses in CoPoNeRF, which provides strong pose-related supervision. Our method, trained with only images, accurately aligns the target pose and synthesizes the image, even from two input frames with large textureless regions.

\subsection{Ablation Studies}
We conduct ablation studies on the RealEstate10K~\cite{re10k} dataset. Since different components are best evaluated under different protocols, we report results under the Target-aligned Evaluation in~\cref{tab:ablation_align}, and results under the Target-aware Evaluation in~\cref{tab:ablation_predict}. It is worth noting that we initialize camera rotations as identity matrices. Given that the RealEstate10K~\cite{re10k} dataset contains a large number of scenes with minimal camera rotation, the RRA@5 metric yields a seemingly high value of 36.6\% for both the Untrained model and models that fail to converge.

\begin{figure}[ht]
\centering
\begin{minipage}{0.41\textwidth}
  \centering
  \includegraphics[width=\linewidth]{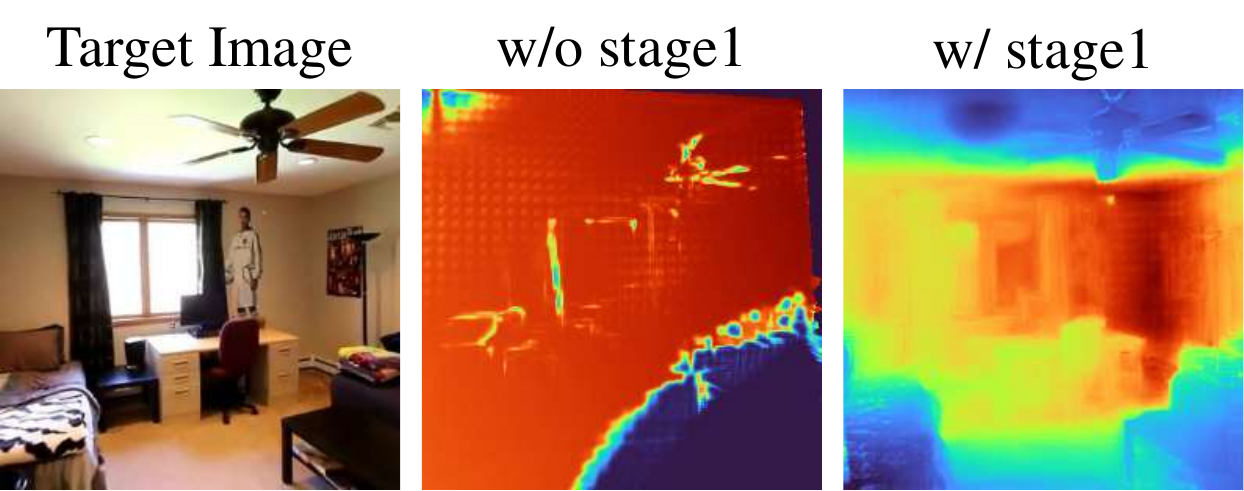}
  \captionsetup{font=small}
  \caption{\small Comparison of the rendered depth maps with and without Stage 1 pretraining.}
  \label{fig:ablation_depth}
\end{minipage}%
\hfill
\begin{minipage}{0.55\textwidth}
  \centering
  \includegraphics[width=\linewidth]{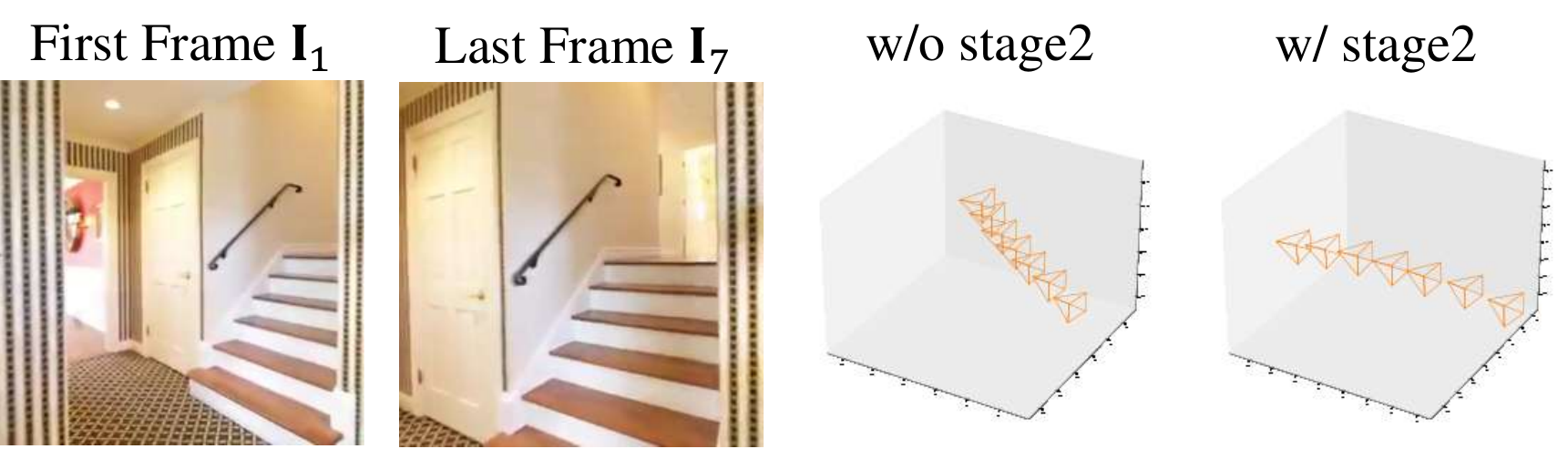}
  \captionsetup{font=small}
  \caption{\small Comparison of the camera trajectories with and without Stage 2 alignment.}
  \label{fig:ablation_pose}
  \end{minipage}
\end{figure}

In~\cref{tab:ablation_align}, the result of w/o Stage 1 shows that our method fails to converge without the implicit reconstruction pretraining. Our proposed implicit reconstruction stage avoids the optimization difficulties of explicit representations and enables the model to learn underlying matching, thereby facilitating convergence. IF refers to the Interpolated Frame Enhanced Prediction described in~\cref{sec:if}. By naturally interpolating frames through interpolated camera poses, our method mitigates the under-convergence issue of the camera network when using only two input views, thus improving performance. \cref{fig:ablation_depth} shows that w/o Stage 1, the model fails to learn meaningful Gaussian primitives.

\Cref{tab:ablation_predict} presents the impact of different training strategies. Consistent with earlier observations, the w/o Stage 1 result again highlights that our stage 1 pretraining facilitates network convergence. The w/o Stage 2 result demonstrates that our network is capable of effectively implicitly reconstructing the scene and using the latent camera representation to predict the target image. Interestingly, due to our use of a physically meaningful Plücker embedding as the camera representation, the predicted latent camera rotations closely approximate the ground-truth rotations, leading to an improved RRA@5 score compared with the untrained model. Qualitatively, we also observe that the latent camera translations tend to align with the ground-truth translations, as shown in~\cref{fig:ablation_pose}. After applying explicit reconstruction alignment, the Full setting demonstrates accurate camera pose estimation, highlighting the effectiveness of our self-supervised alignment strategy. Meanwhile, the image synthesis quality shows a slight degradation due to enforced 3D consistency.

\begin{table}[t!]
\centering
\vspace{-\baselineskip}
\begin{minipage}{0.485\textwidth}
\centering
\captionsetup{font=small}
\caption{\small Target-aligned Ablation Studies.
% IF refers to the Interpolated Frame Enhanced Prediction proposed in~\cref{sec:if}. 
% The results show that our stage1 pretraining facilitates network convergence, while the IF effectively mitigates the issue of insufficient camera alignment under two-frame input settings.
}
\label{tab:ablation_align}

\resizebox{1\linewidth}{!}{
\begin{tabular}{cccccc}
\toprule

Method & PSNR↑ & SSIM↑ & LPIPS↓ & RRA@5↑ & RTA@5↑   \\
\hline 
w/o Stage 1 & 12.97 & 0.401 & 0.529 & 36.6 & 2.28\\
w/o IF & 23.25 & 0.752 & 0.163 & 78.5 & 41.0 \\
Full & \textbf{23.96} & \textbf{0.778} & \textbf{0.145} & \textbf{85.5} & \textbf{50.3} \\

\bottomrule

\end{tabular}
}
\end{minipage}%
\hfill
\begin{minipage}{0.485\textwidth}
\centering
\captionsetup{font=small}
\caption{\small Target-aware ablation. 
% Without Stage 2, the network can reconstruct target images using the predicted latent camera. Our full method improves pose accuracy, with a slight drop in image quality due to enforced 3D consistency.
}
\label{tab:ablation_predict}

\resizebox{1\linewidth}{!}{
\begin{tabular}{cccccc}
\toprule

Method & PSNR↑ & SSIM↑ & LPIPS↓ & RRA@5↑ & RTA@5↑   \\
\hline 
Untrained & 10.66 & 0.037 & 0.830 & 36.6 & 0.00 \\
w/o Stage 1 & 15.71 & 0.479 & 0.448 & 36.6 & 2.36 \\
w/o Stage 2 & \textbf{27.30} & \textbf{0.858} & \textbf{0.107} & 64.2 & 0.10 \\
Full & 26.62 & 0.846 & 0.113 & \textbf{88.2} & \textbf{53.1} \\

\bottomrule

\end{tabular}
}
\end{minipage}
\vspace{-\baselineskip}
\end{table}

\section{Conclusion}
We propose a two-stage training strategy for learning novel view synthesis models from uncalibrated videos. In the first stage, we perform implicit reconstruction pretraining, which addresses the optimization limitations of explicit 3D representations and enables the network to implicitly learn better correspondences. However, since the latent space is not necessarily aligned with the real 3D world, it limits the model to performing novel view synthesis from any given camera. To address this, we introduce explicit reconstruction alignment in the second stage. Specifically, we predict explicit Gaussian primitives and additionally compute the 3DGS rendering loss and depth reprojection loss, injecting 3D consistency in a fully self-supervised manner. To further alleviate the issue of insufficient camera alignment when using only two input frames, we propose an interpolation-based prediction strategy. Experimental results demonstrate that our pretraining phase facilitates network convergence, while the second stage effectively aligns the latent reconstruction with the real 3D space, resulting in high-quality novel view synthesis and accurate camera pose estimation.
\vspace{-\baselineskip}
\paragraph{Limitations.} We assume that input videos depict static scenes, and thus our method is not applicable to the reconstruction of dynamic environments, which remains an interesting future direction.
%Our approach does not involve generative capability and is evaluated solely on publicly available datasets. When extending the method to generative tasks or larger datasets, it is important to ensure data integrity and content safety. When deployed in high-precision systems, the reconstruction quality should be carefully evaluated and further improved if necessary.

\medskip

{
\small
\bibliographystyle{ieeenat_fullname}
\bibliography{pensieve}

\begin{thebibliography}{60}
\providecommand{\natexlab}[1]{#1}
\providecommand{\url}[1]{\texttt{#1}}
\expandafter\ifx\csname urlstyle\endcsname\relax
  \providecommand{\doi}[1]{doi: #1}\else
  \providecommand{\doi}{doi: \begingroup \urlstyle{rm}\Url}\fi

\bibitem[Agarwal et~al.(2011)Agarwal, Furukawa, Snavely, Simon, Curless, Seitz, and Szeliski]{buildingrome}
Sameer Agarwal, Yasutaka Furukawa, Noah Snavely, Ian Simon, Brian Curless, Steven~M Seitz, and Richard Szeliski.
\newblock Building rome in a day.
\newblock \emph{Communications of the ACM}, 54\penalty0 (10):\penalty0 105--112, 2011.

\bibitem[Barron et~al.(2021)Barron, Mildenhall, Tancik, Hedman, Martin-Brualla, and Srinivasan]{mipnerf}
Jonathan~T Barron, Ben Mildenhall, Matthew Tancik, Peter Hedman, Ricardo Martin-Brualla, and Pratul~P Srinivasan.
\newblock Mip-nerf: A multiscale representation for anti-aliasing neural radiance fields.
\newblock In \emph{Proceedings of the IEEE/CVF international conference on computer vision}, pages 5855--5864, 2021.

\bibitem[Bian et~al.(2023)Bian, Wang, Li, Bian, and Prisacariu]{noponerf}
Wenjing Bian, Zirui Wang, Kejie Li, Jia-Wang Bian, and Victor~Adrian Prisacariu.
\newblock Nope-nerf: Optimising neural radiance field with no pose prior.
\newblock In \emph{Proceedings of the IEEE/CVF Conference on Computer Vision and Pattern Recognition}, pages 4160--4169, 2023.

\bibitem[Charatan et~al.(2024)Charatan, Li, Tagliasacchi, and Sitzmann]{pixelsplat}
David Charatan, Sizhe~Lester Li, Andrea Tagliasacchi, and Vincent Sitzmann.
\newblock pixelsplat: 3d gaussian splats from image pairs for scalable generalizable 3d reconstruction.
\newblock In \emph{Proceedings of the IEEE/CVF conference on computer vision and pattern recognition}, pages 19457--19467, 2024.

\bibitem[Chen et~al.(2021)Chen, Xu, Zhao, Zhang, Xiang, Yu, and Su]{chen2021mvsnerf}
Anpei Chen, Zexiang Xu, Fuqiang Zhao, Xiaoshuai Zhang, Fanbo Xiang, Jingyi Yu, and Hao Su.
\newblock Mvsnerf: Fast generalizable radiance field reconstruction from multi-view stereo.
\newblock In \emph{Proceedings of the IEEE/CVF international conference on computer vision}, pages 14124--14133, 2021.

\bibitem[Chen et~al.(2022)Chen, Xu, Geiger, Yu, and Su]{chen2022tensorf}
Anpei Chen, Zexiang Xu, Andreas Geiger, Jingyi Yu, and Hao Su.
\newblock Tensorf: Tensorial radiance fields.
\newblock In \emph{European conference on computer vision}, pages 333--350. Springer, 2022.

\bibitem[Chen et~al.(2024)Chen, Xu, Zheng, Zhuang, Pollefeys, Geiger, Cham, and Cai]{mvsplat}
Yuedong Chen, Haofei Xu, Chuanxia Zheng, Bohan Zhuang, Marc Pollefeys, Andreas Geiger, Tat-Jen Cham, and Jianfei Cai.
\newblock Mvsplat: Efficient 3d gaussian splatting from sparse multi-view images.
\newblock In \emph{European Conference on Computer Vision}, pages 370--386. Springer, 2024.

\bibitem[Fridovich-Keil et~al.(2023)Fridovich-Keil, Meanti, Warburg, Recht, and Kanazawa]{kplanes}
Sara Fridovich-Keil, Giacomo Meanti, Frederik~Rahb{\ae}k Warburg, Benjamin Recht, and Angjoo Kanazawa.
\newblock K-planes: Explicit radiance fields in space, time, and appearance.
\newblock In \emph{Proceedings of the IEEE/CVF Conference on Computer Vision and Pattern Recognition}, pages 12479--12488, 2023.

\bibitem[Fu et~al.(2024)Fu, Liu, Kulkarni, Kautz, Efros, and Wang]{colmapfree3dgs}
Yang Fu, Sifei Liu, Amey Kulkarni, Jan Kautz, Alexei~A Efros, and Xiaolong Wang.
\newblock Colmap-free 3d gaussian splatting.
\newblock In \emph{Proceedings of the IEEE/CVF Conference on Computer Vision and Pattern Recognition}, pages 20796--20805, 2024.

\bibitem[Godard et~al.(2017)Godard, Mac~Aodha, and Brostow]{monodepth1}
Cl{\'e}ment Godard, Oisin Mac~Aodha, and Gabriel~J Brostow.
\newblock Unsupervised monocular depth estimation with left-right consistency.
\newblock In \emph{Proceedings of the IEEE conference on computer vision and pattern recognition}, pages 270--279, 2017.

\bibitem[Godard et~al.(2019)Godard, Mac~Aodha, Firman, and Brostow]{monodepth2}
Cl{\'e}ment Godard, Oisin Mac~Aodha, Michael Firman, and Gabriel~J Brostow.
\newblock Digging into self-supervised monocular depth estimation.
\newblock In \emph{Proceedings of the IEEE/CVF international conference on computer vision}, pages 3828--3838, 2019.

\bibitem[GonzalezBello and Kim(2020)]{falnet}
Juan~Luis GonzalezBello and Munchurl Kim.
\newblock Forget about the lidar: Self-supervised depth estimators with med probability volumes.
\newblock \emph{Advances in Neural Information Processing Systems}, 33:\penalty0 12626--12637, 2020.

\bibitem[Hong et~al.(2024{\natexlab{a}})Hong, Jung, Shin, Han, Yang, Luo, and Kim]{hong2024pf3plat}
Sunghwan Hong, Jaewoo Jung, Heeseong Shin, Jisang Han, Jiaolong Yang, Chong Luo, and Seungryong Kim.
\newblock Pf3plat: Pose-free feed-forward 3d gaussian splatting.
\newblock \emph{arXiv preprint arXiv:2410.22128}, 2024{\natexlab{a}}.

\bibitem[Hong et~al.(2024{\natexlab{b}})Hong, Jung, Shin, Yang, Kim, and Luo]{coponerf}
Sunghwan Hong, Jaewoo Jung, Heeseong Shin, Jiaolong Yang, Seungryong Kim, and Chong Luo.
\newblock Unifying correspondence pose and nerf for generalized pose-free novel view synthesis.
\newblock In \emph{Proceedings of the IEEE/CVF Conference on Computer Vision and Pattern Recognition}, pages 20196--20206, 2024{\natexlab{b}}.

\bibitem[Huang et~al.(2024{\natexlab{a}})Huang, Yu, Chen, Geiger, and Gao]{2dgs}
Binbin Huang, Zehao Yu, Anpei Chen, Andreas Geiger, and Shenghua Gao.
\newblock 2d gaussian splatting for geometrically accurate radiance fields.
\newblock In \emph{ACM SIGGRAPH 2024 conference papers}, pages 1--11, 2024{\natexlab{a}}.

\bibitem[Huang et~al.(2024{\natexlab{b}})Huang, Sun, Yang, Lyu, Cao, and Qi]{huang2024scgs}
Yi-Hua Huang, Yang-Tian Sun, Ziyi Yang, Xiaoyang Lyu, Yan-Pei Cao, and Xiaojuan Qi.
\newblock Sc-gs: Sparse-controlled gaussian splatting for editable dynamic scenes.
\newblock In \emph{Proceedings of the IEEE/CVF conference on computer vision and pattern recognition}, pages 4220--4230, 2024{\natexlab{b}}.

\bibitem[Jeong et~al.(2021)Jeong, Ahn, Choy, Anandkumar, Cho, and Park]{scnerf}
Yoonwoo Jeong, Seokjun Ahn, Christopher Choy, Anima Anandkumar, Minsu Cho, and Jaesik Park.
\newblock Self-calibrating neural radiance fields.
\newblock In \emph{Proceedings of the IEEE/CVF International Conference on Computer Vision}, pages 5846--5854, 2021.

\bibitem[Jiang et~al.(2024)Jiang, Fu, Varma~T, Belhe, Wang, Su, and Ramamoorthi]{jiang2024construct}
Kaiwen Jiang, Yang Fu, Mukund Varma~T, Yash Belhe, Xiaolong Wang, Hao Su, and Ravi Ramamoorthi.
\newblock A construct-optimize approach to sparse view synthesis without camera pose.
\newblock In \emph{ACM SIGGRAPH 2024 Conference Papers}, pages 1--11, 2024.

\bibitem[Jin et~al.(2025)Jin, Jiang, Tan, Zhang, Bi, Zhang, Luan, Snavely, and Xu]{lvsm}
Haian Jin, Hanwen Jiang, Hao Tan, Kai Zhang, Sai Bi, Tianyuan Zhang, Fujun Luan, Noah Snavely, and Zexiang Xu.
\newblock Lvsm: A large view synthesis model with minimal 3d inductive bias.
\newblock In \emph{The Thirteenth International Conference on Learning Representations}, 2025.

\bibitem[Kang et~al.(2024)Kang, Yoo, Park, Nam, Im, Shin, Kim, and Park]{selfsplat}
Gyeongjin Kang, Jisang Yoo, Jihyeon Park, Seungtae Nam, Hyeonsoo Im, Sangheon Shin, Sangpil Kim, and Eunbyung Park.
\newblock Selfsplat: Pose-free and 3d prior-free generalizable 3d gaussian splatting.
\newblock \emph{arXiv preprint arXiv:2411.17190}, 2024.

\bibitem[Kerbl et~al.(2023)Kerbl, Kopanas, Leimk{\"u}hler, and Drettakis]{3dgs}
Bernhard Kerbl, Georgios Kopanas, Thomas Leimk{\"u}hler, and George Drettakis.
\newblock 3d gaussian splatting for real-time radiance field rendering.
\newblock \emph{ACM Trans. Graph.}, 42\penalty0 (4):\penalty0 139--1, 2023.

\bibitem[Kerl et~al.(2013)Kerl, Sturm, and Cremers]{densergbdslam}
Christian Kerl, J{\"u}rgen Sturm, and Daniel Cremers.
\newblock Dense visual slam for rgb-d cameras.
\newblock In \emph{2013 IEEE/RSJ international conference on intelligent robots and systems}, pages 2100--2106. IEEE, 2013.

\bibitem[Leroy et~al.(2024)Leroy, Cabon, and Revaud]{mast3r}
Vincent Leroy, Yohann Cabon, and J{\'e}r{\^o}me Revaud.
\newblock Grounding image matching in 3d with mast3r.
\newblock In \emph{European Conference on Computer Vision}, pages 71--91. Springer, 2024.

\bibitem[Li et~al.(2024)Li, Gao, Wu, Zhang, Dai, Zhao, Feng, Ding, Wang, and Han]{li2024ggrt}
Hao Li, Yuanyuan Gao, Chenming Wu, Dingwen Zhang, Yalun Dai, Chen Zhao, Haocheng Feng, Errui Ding, Jingdong Wang, and Junwei Han.
\newblock Ggrt: Towards pose-free generalizable 3d gaussian splatting in real-time.
\newblock In \emph{European Conference on Computer Vision}, pages 325--341. Springer, 2024.

\bibitem[Li et~al.(2021{\natexlab{a}})Li, Feng, She, Ding, Wang, and Lee]{li2021mine}
Jiaxin Li, Zijian Feng, Qi She, Henghui Ding, Changhu Wang, and Gim~Hee Lee.
\newblock Mine: Towards continuous depth mpi with nerf for novel view synthesis.
\newblock In \emph{Proceedings of the IEEE/CVF International Conference on Computer Vision}, pages 12578--12588, 2021{\natexlab{a}}.

\bibitem[Li et~al.(2021{\natexlab{b}})Li, Niklaus, Snavely, and Wang]{nerfsceneflow}
Zhengqi Li, Simon Niklaus, Noah Snavely, and Oliver Wang.
\newblock Neural scene flow fields for space-time view synthesis of dynamic scenes.
\newblock In \emph{Proceedings of the IEEE/CVF Conference on Computer Vision and Pattern Recognition}, pages 6498--6508, 2021{\natexlab{b}}.

\bibitem[Li et~al.(2023)Li, M{\"u}ller, Evans, Taylor, Unberath, Liu, and Lin]{li2023neuralangelo}
Zhaoshuo Li, Thomas M{\"u}ller, Alex Evans, Russell~H Taylor, Mathias Unberath, Ming-Yu Liu, and Chen-Hsuan Lin.
\newblock Neuralangelo: High-fidelity neural surface reconstruction.
\newblock In \emph{Proceedings of the IEEE/CVF Conference on Computer Vision and Pattern Recognition}, pages 8456--8465, 2023.

\bibitem[Li et~al.(2025)Li, Dong, Chen, Huang, and Liu]{li2025vicasplat}
Zhiqi Li, Chengrui Dong, Yiming Chen, Zhangchi Huang, and Peidong Liu.
\newblock Vicasplat: A single run is all you need for 3d gaussian splatting and camera estimation from unposed video frames.
\newblock \emph{arXiv preprint arXiv:2503.10286}, 2025.

\bibitem[Lin et~al.(2021)Lin, Ma, Torralba, and Lucey]{barf}
Chen-Hsuan Lin, Wei-Chiu Ma, Antonio Torralba, and Simon Lucey.
\newblock Barf: Bundle-adjusting neural radiance fields.
\newblock In \emph{Proceedings of the IEEE/CVF international conference on computer vision}, pages 5741--5751, 2021.

\bibitem[Lindenberger et~al.(2023)Lindenberger, Sarlin, and Pollefeys]{lindenberger2023lightglue}
Philipp Lindenberger, Paul-Edouard Sarlin, and Marc Pollefeys.
\newblock Lightglue: Local feature matching at light speed.
\newblock In \emph{Proceedings of the IEEE/CVF International Conference on Computer Vision}, pages 17627--17638, 2023.

\bibitem[Ling et~al.(2024)Ling, Sheng, Tu, Zhao, Xin, Wan, Yu, Guo, Yu, Lu, et~al.]{ling2024dl3dv}
Lu Ling, Yichen Sheng, Zhi Tu, Wentian Zhao, Cheng Xin, Kun Wan, Lantao Yu, Qianyu Guo, Zixun Yu, Yawen Lu, et~al.
\newblock Dl3dv-10k: A large-scale scene dataset for deep learning-based 3d vision.
\newblock In \emph{Proceedings of the IEEE/CVF Conference on Computer Vision and Pattern Recognition}, pages 22160--22169, 2024.

\bibitem[Lu et~al.(2024)Lu, Yu, Xu, Xiangli, Wang, Lin, and Dai]{lu2024scaffold}
Tao Lu, Mulin Yu, Linning Xu, Yuanbo Xiangli, Limin Wang, Dahua Lin, and Bo Dai.
\newblock Scaffold-gs: Structured 3d gaussians for view-adaptive rendering.
\newblock In \emph{Proceedings of the IEEE/CVF Conference on Computer Vision and Pattern Recognition}, pages 20654--20664, 2024.

\bibitem[Matsuki et~al.(2024)Matsuki, Murai, Kelly, and Davison]{monogs}
Hidenobu Matsuki, Riku Murai, Paul~HJ Kelly, and Andrew~J Davison.
\newblock Gaussian splatting slam.
\newblock In \emph{Proceedings of the IEEE/CVF Conference on Computer Vision and Pattern Recognition}, pages 18039--18048, 2024.

\bibitem[Meuleman et~al.(2023)Meuleman, Liu, Gao, Huang, Kim, Kim, and Kopf]{localrf}
Andreas Meuleman, Yu-Lun Liu, Chen Gao, Jia-Bin Huang, Changil Kim, Min~H Kim, and Johannes Kopf.
\newblock Progressively optimized local radiance fields for robust view synthesis.
\newblock In \emph{Proceedings of the IEEE/CVF Conference on Computer Vision and Pattern Recognition}, pages 16539--16548, 2023.

\bibitem[Mildenhall et~al.(2021)Mildenhall, Srinivasan, Tancik, Barron, Ramamoorthi, and Ng]{nerf}
Ben Mildenhall, Pratul~P Srinivasan, Matthew Tancik, Jonathan~T Barron, Ravi Ramamoorthi, and Ren Ng.
\newblock Nerf: Representing scenes as neural radiance fields for view synthesis.
\newblock \emph{Communications of the ACM}, 65\penalty0 (1):\penalty0 99--106, 2021.

\bibitem[M{\"u}ller et~al.(2022)M{\"u}ller, Evans, Schied, and Keller]{ingp}
Thomas M{\"u}ller, Alex Evans, Christoph Schied, and Alexander Keller.
\newblock Instant neural graphics primitives with a multiresolution hash encoding.
\newblock \emph{ACM transactions on graphics (TOG)}, 41\penalty0 (4):\penalty0 1--15, 2022.

\bibitem[Park et~al.(2024)Park, Bui, Bello, Moon, Oh, and Kim]{park2024splinegs}
Jongmin Park, Minh-Quan~Viet Bui, Juan Luis~Gonzalez Bello, Jaeho Moon, Jihyong Oh, and Munchurl Kim.
\newblock Splinegs: Robust motion-adaptive spline for real-time dynamic 3d gaussians from monocular video.
\newblock \emph{arXiv preprint arXiv:2412.09982}, 2024.

\bibitem[Park et~al.(2021)Park, Sinha, Barron, Bouaziz, Goldman, Seitz, and Martin-Brualla]{park2021nerfies}
Keunhong Park, Utkarsh Sinha, Jonathan~T Barron, Sofien Bouaziz, Dan~B Goldman, Steven~M Seitz, and Ricardo Martin-Brualla.
\newblock Nerfies: Deformable neural radiance fields.
\newblock In \emph{Proceedings of the IEEE/CVF international conference on computer vision}, pages 5865--5874, 2021.

\bibitem[Sch\"{o}nberger and Frahm(2016)]{colmap}
Johannes~Lutz Sch\"{o}nberger and Jan-Michael Frahm.
\newblock Structure-from-motion revisited.
\newblock In \emph{Conference on Computer Vision and Pattern Recognition (CVPR)}, 2016.

\bibitem[Smart et~al.(2024)Smart, Zheng, Laina, and Prisacariu]{smart2024splatt3r}
Brandon Smart, Chuanxia Zheng, Iro Laina, and Victor~Adrian Prisacariu.
\newblock Splatt3r: Zero-shot gaussian splatting from uncalibrated image pairs.
\newblock \emph{arXiv preprint arXiv:2408.13912}, 2024.

\bibitem[Smith et~al.(2023)Smith, Du, Tewari, and Sitzmann]{smith2023flowcam}
Cameron Smith, Yilun Du, Ayush Tewari, and Vincent Sitzmann.
\newblock Flowcam: Training generalizable 3d radiance fields without camera poses via pixel-aligned scene flow.
\newblock \emph{arXiv preprint arXiv:2306.00180}, 2023.

\bibitem[Suhail et~al.(2022)Suhail, Esteves, Sigal, and Makadia]{suhail2022light}
Mohammed Suhail, Carlos Esteves, Leonid Sigal, and Ameesh Makadia.
\newblock Light field neural rendering.
\newblock In \emph{Proceedings of the IEEE/CVF Conference on Computer Vision and Pattern Recognition}, pages 8269--8279, 2022.

\bibitem[Szymanowicz et~al.(2024)Szymanowicz, Rupprecht, and Vedaldi]{szymanowicz2024splatter}
Stanislaw Szymanowicz, Chrisitian Rupprecht, and Andrea Vedaldi.
\newblock Splatter image: Ultra-fast single-view 3d reconstruction.
\newblock In \emph{Proceedings of the IEEE/CVF conference on computer vision and pattern recognition}, pages 10208--10217, 2024.

\bibitem[Tang et~al.(2024)Tang, Chen, Chen, Wang, Zeng, and Liu]{tang2024lgm}
Jiaxiang Tang, Zhaoxi Chen, Xiaokang Chen, Tengfei Wang, Gang Zeng, and Ziwei Liu.
\newblock Lgm: Large multi-view gaussian model for high-resolution 3d content creation.
\newblock In \emph{European Conference on Computer Vision}, pages 1--18. Springer, 2024.

\bibitem[Tian et~al.(2023)Tian, Du, and Duan]{tian2023mononerf}
Fengrui Tian, Shaoyi Du, and Yueqi Duan.
\newblock Mononerf: Learning a generalizable dynamic radiance field from monocular videos.
\newblock In \emph{Proceedings of the IEEE/CVF International Conference on Computer Vision}, pages 17903--17913, 2023.

\bibitem[Truong et~al.(2023)Truong, Rakotosaona, Manhardt, and Tombari]{truong2023sparf}
Prune Truong, Marie-Julie Rakotosaona, Fabian Manhardt, and Federico Tombari.
\newblock Sparf: Neural radiance fields from sparse and noisy poses.
\newblock In \emph{Proceedings of the IEEE/CVF Conference on Computer Vision and Pattern Recognition}, pages 4190--4200, 2023.

\bibitem[Wang et~al.(2021{\natexlab{a}})Wang, Wang, Genova, Srinivasan, Zhou, Barron, Martin-Brualla, Snavely, and Funkhouser]{wang2021ibrnet}
Qianqian Wang, Zhicheng Wang, Kyle Genova, Pratul~P Srinivasan, Howard Zhou, Jonathan~T Barron, Ricardo Martin-Brualla, Noah Snavely, and Thomas Funkhouser.
\newblock Ibrnet: Learning multi-view image-based rendering.
\newblock In \emph{Proceedings of the IEEE/CVF conference on computer vision and pattern recognition}, pages 4690--4699, 2021{\natexlab{a}}.

\bibitem[Wang et~al.(2023)Wang, Yu, and Gao]{planedepth}
Ruoyu Wang, Zehao Yu, and Shenghua Gao.
\newblock Planedepth: Self-supervised depth estimation via orthogonal planes.
\newblock In \emph{Proceedings of the IEEE/CVF Conference on Computer Vision and Pattern Recognition}, pages 21425--21434, 2023.

\bibitem[Wang et~al.(2024)Wang, Leroy, Cabon, Chidlovskii, and Revaud]{wang2024dust3r}
Shuzhe Wang, Vincent Leroy, Yohann Cabon, Boris Chidlovskii, and Jerome Revaud.
\newblock Dust3r: Geometric 3d vision made easy.
\newblock In \emph{Proceedings of the IEEE/CVF Conference on Computer Vision and Pattern Recognition}, pages 20697--20709, 2024.

\bibitem[Wang et~al.(2021{\natexlab{b}})Wang, Wu, Xie, Chen, and Prisacariu]{nerfmm}
Zirui Wang, Shangzhe Wu, Weidi Xie, Min Chen, and Victor~Adrian Prisacariu.
\newblock Nerf--: Neural radiance fields without known camera parameters.
\newblock \emph{arXiv preprint arXiv:2102.07064}, 2021{\natexlab{b}}.

\bibitem[Xu et~al.(2024)Xu, Shi, Yifan, Chen, Yang, Peng, Shen, and Wetzstein]{xu2024grm}
Yinghao Xu, Zifan Shi, Wang Yifan, Hansheng Chen, Ceyuan Yang, Sida Peng, Yujun Shen, and Gordon Wetzstein.
\newblock Grm: Large gaussian reconstruction model for efficient 3d reconstruction and generation.
\newblock In \emph{European Conference on Computer Vision}, pages 1--20. Springer, 2024.

\bibitem[Yang et~al.(2024)Yang, Zhu, Jiang, Ye, Chen, Zhang, Chen, Zhao, and Zhao]{yang2024spectrally}
Runyi Yang, Zhenxin Zhu, Zhou Jiang, Baijun Ye, Xiaoxue Chen, Yifei Zhang, Yuantao Chen, Jian Zhao, and Hao Zhao.
\newblock Spectrally pruned gaussian fields with neural compensation.
\newblock \emph{arXiv preprint arXiv:2405.00676}, 2024.

\bibitem[Ye et~al.(2024)Ye, Liu, Xu, Li, Pollefeys, Yang, and Peng]{noposplat}
Botao Ye, Sifei Liu, Haofei Xu, Xueting Li, Marc Pollefeys, Ming-Hsuan Yang, and Songyou Peng.
\newblock No pose, no problem: Surprisingly simple 3d gaussian splats from sparse unposed images.
\newblock \emph{arXiv preprint arXiv:2410.24207}, 2024.

\bibitem[Yu et~al.(2021)Yu, Ye, Tancik, and Kanazawa]{yu2021pixelnerf}
Alex Yu, Vickie Ye, Matthew Tancik, and Angjoo Kanazawa.
\newblock pixelnerf: Neural radiance fields from one or few images.
\newblock In \emph{Proceedings of the IEEE/CVF conference on computer vision and pattern recognition}, pages 4578--4587, 2021.

\bibitem[Yu et~al.(2024{\natexlab{a}})Yu, Julin, Milacski, Niinuma, and Jeni]{yu2024cogs}
Heng Yu, Joel Julin, Zolt{\'a}n~{\'A} Milacski, Koichiro Niinuma, and L{\'a}szl{\'o}~A Jeni.
\newblock Cogs: Controllable gaussian splatting.
\newblock In \emph{Proceedings of the IEEE/CVF Conference on Computer Vision and Pattern Recognition}, pages 21624--21633, 2024{\natexlab{a}}.

\bibitem[Yu et~al.(2024{\natexlab{b}})Yu, Chen, Huang, Sattler, and Geiger]{mipsplatting}
Zehao Yu, Anpei Chen, Binbin Huang, Torsten Sattler, and Andreas Geiger.
\newblock Mip-splatting: Alias-free 3d gaussian splatting.
\newblock In \emph{Proceedings of the IEEE/CVF conference on computer vision and pattern recognition}, pages 19447--19456, 2024{\natexlab{b}}.

\bibitem[Zhang et~al.(2024)Zhang, Bi, Tan, Xiangli, Zhao, Sunkavalli, and Xu]{gslrm}
Kai Zhang, Sai Bi, Hao Tan, Yuanbo Xiangli, Nanxuan Zhao, Kalyan Sunkavalli, and Zexiang Xu.
\newblock Gs-lrm: Large reconstruction model for 3d gaussian splatting.
\newblock In \emph{European Conference on Computer Vision}, pages 1--19. Springer, 2024.

\bibitem[Zhang et~al.(2025)Zhang, Wang, Xu, Xue, Rupprecht, Zhou, Shen, and Wetzstein]{zhang2025flare}
Shangzhan Zhang, Jianyuan Wang, Yinghao Xu, Nan Xue, Christian Rupprecht, Xiaowei Zhou, Yujun Shen, and Gordon Wetzstein.
\newblock Flare: Feed-forward geometry, appearance and camera estimation from uncalibrated sparse views.
\newblock \emph{arXiv preprint arXiv:2502.12138}, 2025.

\bibitem[Zhou et~al.(2017)Zhou, Brown, Snavely, and Lowe]{sfmlearner}
Tinghui Zhou, Matthew Brown, Noah Snavely, and David~G Lowe.
\newblock Unsupervised learning of depth and ego-motion from video.
\newblock In \emph{Proceedings of the IEEE conference on computer vision and pattern recognition}, pages 1851--1858, 2017.

\bibitem[Zhou et~al.(2018)Zhou, Tucker, Flynn, Fyffe, and Snavely]{re10k}
Tinghui Zhou, Richard Tucker, John Flynn, Graham Fyffe, and Noah Snavely.
\newblock Stereo magnification: Learning view synthesis using multiplane images.
\newblock \emph{ACM Trans. Graph. (Proc. SIGGRAPH)}, 37, 2018.

\end{thebibliography}
}

%%%%%%%%%%%%%%%%%%%%%%%%%%%%%%%%%%%%%%%%%%%%%%%%%%%%%%%%%%%%

% \appendix

% \section{Technical Appendices and Supplementary Material}
% Technical appendices with additional results, figures, graphs and proofs may be submitted with the paper submission before the full submission deadline (see above), or as a separate PDF in the ZIP file below before the supplementary material deadline. There is no page limit for the technical appendices.

%%%%%%%%%%%%%%%%%%%%%%%%%%%%%%%%%%%%%%%%%%%%%%%%%%%%%%%%%%%%

% \input{sec/checklist}

\end{document}